%% file: main.tex
\documentclass[acmsmall, nonacm, screen]{acmart}
\usepackage{amsmath,amsthm,color}

\newcommand{\cI}{\mathcal{I}}   
\newcommand{\cJ}{\mathcal{J}}   
   
\newcommand{\cG}{\mathcal{G}}   
\newcommand{\cT}{\mathcal{T}}   
\newcommand{\R}{\mathbb{R}}   

\AtBeginDocument{%
  \providecommand\BibTeX{{%
    \normalfont B\kern-0.5em{\scshape i\kern-0.25em b}\kern-0.8em\TeX}}}

\begin{document}

\title{FaceFolds: Meshed Radiance Manifolds for Efficient Volumetric Rendering of Dynamic Faces}
\author{Safa C. Medin}
\email{medin@mit.edu}
\affiliation{%
  \institution{MIT and Google}
  \country{USA}
}

\author{Gengyan Li}
\email{gengyan.li@inf.ethz.ch}
\affiliation{%
  \institution{ETH Zurich and Google}
  \country{Switzerland}}

\author{Ruofei Du}
\email{me@duruofei.com}
\affiliation{%
  \institution{Google}
  \country{USA}}
  
\author{Stephan Garbin}
\email{stephangarbin@google.com}
\affiliation{%
  \institution{Google}
  \country{United Kingdom}}

\author{Philip Davidson}
\email{pdavidson@google.com}
\affiliation{%
  \institution{Google}
  \country{USA}}

\author{Gregory W. Wornell}
\email{gww@mit.edu}
\affiliation{%
  \institution{MIT}
  \country{USA}}
  
\author{Thabo Beeler}
\email{tbeeler@google.com}
\affiliation{%
  \institution{Google}
  \country{Switzerland}}
  
\author{Abhimitra Meka}
\email{abhim@google.com}
\affiliation{%
  \institution{Google}
  \country{USA}}

\renewcommand{\shortauthors}{Medin et al.}

\input{sections/0_abstract} 



\begin{teaserfigure}
\includegraphics[width=\textwidth]{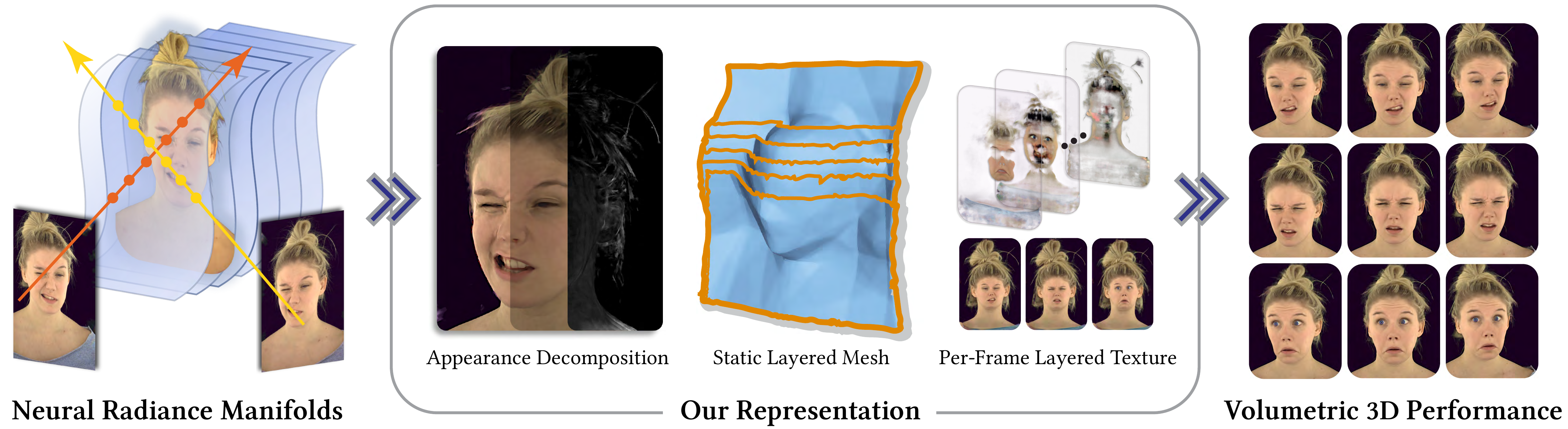}
\caption{We model a dynamic face sequence as a set of radiance manifolds, which are exported as a static layered mesh and an animated texture. This allows for smoothly controlling the quality vs.\ memory and compute footprints, while achieving efficient and photorealistic rendering of volumetric scenes using established graphics pipelines without any neural network integration. \url{https://syntec-research.github.io/FaceFolds/}}
\label{fig:teaser}
\end{teaserfigure}

\maketitle
\input{sections/1_intro}
\input{sections/2_related_work}
\input{sections/3_method}
\input{sections/4_results}
\input{sections/5_future_work}

\input{sections/6_conclusion}
{
    \small
    \bibliographystyle{ACM-Reference-Format}
    \bibliography{main}
}

\end{document}

%% file: sections/0_abstract.tex
\begin{abstract}
3D rendering of dynamic face captures is a challenging problem, and it demands improvements on several fronts---photorealism, efficiency, compatibility, and configurability. We present a novel representation that enables high-quality volumetric rendering of an actor's dynamic facial performances with minimal compute and memory footprint. It runs natively on commodity graphics soft- and hardware, and allows for a graceful trade-off between quality and efficiency. Our method utilizes recent advances in neural rendering, particularly learning discrete radiance manifolds to sparsely sample the scene to model volumetric effects. We achieve efficient modeling by learning a single set of manifolds for the entire dynamic sequence, while implicitly modeling appearance changes as temporal canonical texture. We export a single layered mesh and view-independent RGBA texture video that is compatible with legacy graphics renderers without additional ML integration. We demonstrate our method by rendering dynamic face captures of real actors in a game engine, at comparable photorealism to state-of-the-art neural rendering techniques at previously unseen frame rates.
\end{abstract}

%% file: sections/1_intro.tex
\section{Introduction}
\label{sec:intro}

Facial expressions are our primary means of communication---video streaming of our faces is a frequent part of our daily digital lives. Perceptually lossless and efficient video compression algorithms have enabled this application at consumer scale by achieving compute and memory efficiency. Other key enablers of this ubiquity of video streaming are 1) compatibility of compressed video playback with existing platform infrastructure such as operating systems and web browsers, and 2) easy trade-off of quality vs.\ memory through variable image resolution, allowing for seamless streaming across dynamically varying data bandwidth. But the same cannot be said for streaming and playback of personalized 3D face animation, which is a formidable task with major algorithmic challenges such as 3D reconstruction, animation, and transmission. While photorealism and compute/memory efficiency are obvious underlying challenges to each of these steps, other hurdles include compatibility with legacy software infrastructure and ability to smoothly trade-off quality vs.\ bandwidth. Solving these challenges promises widespread adoption of immersive experiences in 3D media content~\cite{fyffe2011,gotardo2018}, immersive AR/VR communication and 3D telepresence~\cite{Orts-Escolano2016Holoportation,Du2018Montage4D,Du2019Montage4D,chen_authentic}.

Traditional graphics-based acquisition techniques for facial performances reconstructs 3D meshes and texture maps~\cite{Guo2019The,Collet2015} for each individual frame, which can be rendered very efficiently on commodity hardware. However, such traditional mesh-based representations encounter significant challenges in accurately modeling and rendering the fine-scale detailed geometry and complex appearance of hair and skin~\cite{Du2018Montage4D,Guo2019The}, resulting in limited photorealism. On the other hand, recent advances in implicit volumetric representations such as neural radiance fields~\cite{mildenhall2020nerf} and Gaussian splatting~\cite{kerbl3Dgaussians} have enabled high-quality acquisition and photorealistic rendering of dynamic human faces~\cite{park2021nerfies,qian2023gaussianavatars,nv}, including hair and skin at unprecedented quality~\cite{buhler2023preface,saito2023rgca}, without requiring explicit geometry reconstruction. But such techniques require deep ML integration for inference and are not natively compatible with existing graphics rendering platforms such as game engines. They also do not always provide means to trade-off quality with compute or memory efficiency.

We present a layered-mesh based volumetric representation for 3D view-synthesis of dynamic face performances that works efficiently on legacy renderers. At training time, our method takes inspiration from radiance manifolds~\cite{deng2022gram} and models the scene density for the entire sequence using a set of static spatial manifolds of alpha values, and the temporal appearance changes as a time-conditioned UV-mapped radiance over these manifolds. The alpha-manifolds and corresponding temporal UV appearance maps are parameterized by dense neural networks, and the appearance is further decomposed into view-conditioned components. From this trained model, we export the radiance manifolds as a single layered mesh for the entire sequence, and the corresponding view-independent UV-space appearance as RGBA texture maps, encoded as a video. This exported representation is then rendered efficiently through simple alpha-blending of the textured mesh layers in any renderer. Unlike previous methods that cannot change the resolution or quality once trained, our layered mesh representation also allows for trading off the image quality for efficiency through standard operations like mesh decimation and subsampling of texture resolution. Our view-independent texture maps allow for easy rendering through texture look-up without the need for evaluating complex view-dependent reflectance or radiance transfer.

We demonstrate the efficacy of our representation using the Multiface dataset \cite{wuu2022multiface} consisting of multi-view real world captures of face performances of several actors. We qualitatively and quantitatively compare our method with state-of-the-art neural rendering techniques on both image quality and efficiency, and show previously unseen high frame-rate rendering of these sequences on the Unity game engine on a consumer device.

%% file: sections/2_related_work.tex
\section{Related Work}
\label{sec:relatedWork}

Impressive results have been achieved for editing and animation of face images and videos using purely 2D or hybrid approaches \cite{kim2018deep, Zakharov2020FastBN,wen2020audiodvp,buehler2021varitex, medin2022most}, but such methods do not guarantee consistent rendering when changing perspectives, which is crucial for gaming or XR applications.

Traditional 3D performance capture techniques ~\cite{fyffe2011,gotardo2018,Guo2019The,Collet2015} have relied on textured meshes both due to ease of use for playback and editing, and to rely on rasterization for fast rendering. Parametric face models such as 3D morphable face models (3DMMs)~\cite{3dmm,flame, egger20203d} compress the dimensionality of mesh representations and make them differentiable, enabling efficient optimization frameworks that achieve the difficult task of canonical performance capture and playback. However, they suffer from low representational capacity and cannot model high-frequency effects in appearance and geometry. Some recent ML-based approaches build on such surface-based representations, \textit{e.g.}, Neural Head Avatars employs a surface mesh with a dynamic texture~\cite{NHA}, while IMAvatar~\cite{imavatar} opts for an implicit surface. While efficient, these methods need to generate expression-specific appearance via feed-forward neural networks and have issues representing semi-transparent effects such as hair and beards. Adding deferred rendering networks, such as in the single-shot model~\cite{Khakhulin2022ROME} struggles to mitigate these effects and jeopardizes multi-view consistency.

Volumetric representations such as NeRFs parameterize a compressed emission-absorption volume via a multi-layer perceptron (MLP) trained with frequency encoding~\cite{nerf, tancik2020fourfeat}. Since the MLP has to be invoked for decompression at every sample location, inference performance stands out as one of the main limitations of the original work. Subsequently, hybrid representations have replaced the need for an MLP at inference time with (sparse) volumetric grids~\cite{plenoxels,Garbin2021FastNeRFHN, hedman2021snerg, Reiser2023SIGGRAPH, duckworth2023smerf, neuralSparseVoxelFields, eg3d}, explicit geometry~\cite{chen2022mobilenerf, deng2022gram, eyenerf}, and hash grids \cite{mueller2022instant}. Many recent works have exploited such volumetric representations to track or animate head models, but typically with computational requirements far exceeding our target application. Nerfies~\cite{park2021nerfies} and HyperNeRF~\cite{park2021hypernerf} propose a continuous deformation field conditioned on a frame-specific latent code, which enables replay but requires an additional deformation MLP. 

A popular class of methods rely on combining the low-dimensional tracking capabilities of 3DMMs with high representational power of volumetric radiance. GNARF~\cite{gnarf} and Next3D~\cite{next3d} use a tri-plane representation with a form of mesh-based deformation but still require 2D super-resolution modules to generate image at the desired resolution. INSTA~\cite{insta} increases the efficiency of this type of approach by using hash grids~\cite{insta}, while NerSemble replaces the explicit head model with a hash ensemble for increased generality \cite{kirschstein2023nersemble}. Other methods~\cite{garbin2022voltemorph, yuan2022nerfediting, kania2023blendfields} employ tetrahedral fields to directly deform a volumetric representation. MonoAvatar~\cite{monoAvatar} uses a full volumetric NeRF model, where deformation is a function of $K$ nearest points on the driving surface mesh. Ray tracing and the use of MLPs means that these methods are not nearly as fast as rasterization-based techniques. MVP~\cite{mvp} and its generative extension~\cite{chen_authentic} offer a more efficient rendering pipeline by employing volumetric primitives attached to a guide mesh. More recent works have used other alternatives to NeRF-like volumes, such as point based representations~\cite{Zheng2023pointavatar} and Gaussian Splatting~\cite{saito2023rgca, qian2023gaussianavatars} driven by combinations of explicit models like FLAME~\cite{flame} and neural functions. While these models offer improved performance compared to volumetric methods, they still require custom rasterization components and are thus not trivial to deploy in existing software. For all these methods, use of 3DMMs makes these methods challenging to deploy without additional software components including iterative optimizers.

An alternate way to mitigate issues of predefined mesh topology and increase the flexibility of the representation is to use one~\cite{wang2021neus, li2023neuralangelo} or a collection of (implicit) surfaces~\cite{deng2022gram}. In GRAM~\cite{deng2022gram}, the authors propose a discretized volumetric representation using a set of learned non-intersecting implicit surfaces which can be efficiently used for sampling radiance. GRAMInverter~\cite{deng2023learning} and GRAM-HD \cite{Xiang_2023_ICCV} propose high-resolution variants of GRAM, albeit at increased computational cost and model complexity not suitable for real-time applications. BakedAvatar~\cite{bakedavatar} is a concurrently developed avatar animation technique that also uses radiance manifolds, and like previous methods, employs deformation fields conditioned on tracked 3DMM coefficients. However, their model still relies on an ML model to estimate appearance at inference time. During training, our method builds on layered implicit surfaces used in GRAM~\cite{deng2022gram}, but is not generative, focusing on high-quality, per-person specific models instead. We show that a static set of manifolds can model an entire performance sequence, and we offer improved performance by decomposing the radiance into view-dependent and -independent components, which allows us to export our results to a single explicit triangle mesh with video texture. This enables real-time playback in commonly available software packages at low computational cost.

%% file: sections/3_method.tex
\section{Method}
In this section, we first formulate the objective problem and present an overview of our pipeline. After describing how we process our datasets, we elaborate on how we leverage radiance manifolds to learn efficient 3D representations from multi-view videos. Finally, we describe how we export our representation to a single set of textured meshes that can be rendered natively on traditional graphics software while maintaining the rendering quality.
\begin{figure}[t]
  \centering
  \includegraphics[width=\linewidth]{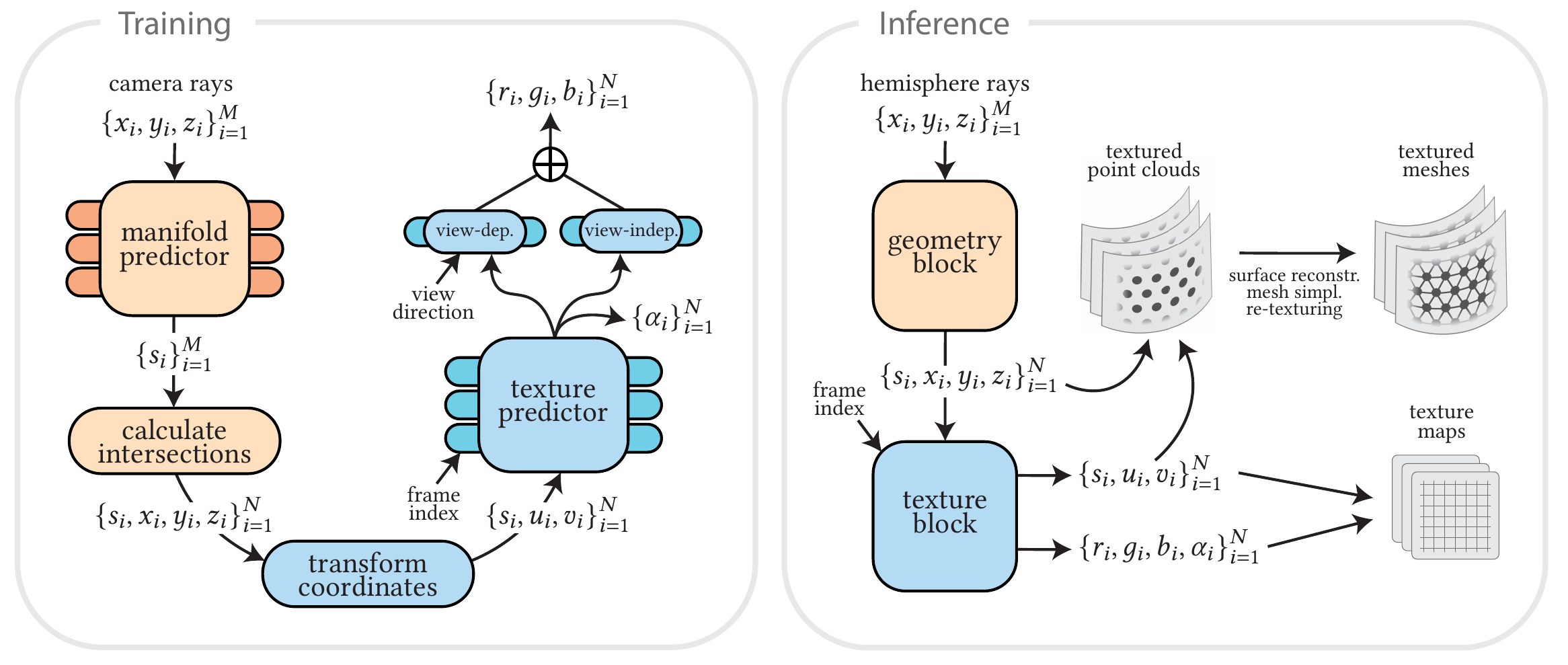}
  \caption{\textbf{Training and inference pipelines.} Given a set of rays from the training cameras, we determine the intersection of these rays with a set of implicit manifolds predicted by a single manifold predictor. After transforming these intersections to UV-space coordinates, a texture predictor outputs RGBA texture maps conditioned on the video frame index. At inference time, we shoot rays from the surface of a designated hemisphere around the scene towards its center, obtaining a single geometry and a video texture. The view-dependent branch is bypassed to ensure that the appearance is fully diffuse.}
  \label{fig:method}
\end{figure}
\label{sec:method}
\subsection{Definitions and Overview}
Our objective is to learn a volumetric 3D representation of a subject that can be played back on game engines without any special neural network integration. Given a multi-view video of a subject with $K$ frames, we learn a static geometry and a dynamic appearance model in an end-to-end fashion. We take inspiration from recent advances in implicit geometry representations \cite{li2023neuralangelo,wang2021neus} that significantly outperform explicit 3D reconstruction techniques that rely on mesh or point cloud representations. In our pipeline, similar to GRAM~\cite{deng2022gram}, the geometry is modeled by a set of 2D manifolds, embodied as a set of implicit surfaces. {But unlike GRAM,} the appearance is learned as a UV-mapped dynamic radiance over these manifolds, instead of the 3D $xyz$-space. We learn $N$ distinct manifolds defined implicitly by a single manifold predictor:
\begin{equation}
    \cG : (x,y,z) \in \R^3 \to s \in \R
\end{equation}
Given a set of fixed scalars $\{s_i \in \R \ | \ i \in \cI \triangleq \{1, 2, \dots, N)\}$, which we refer to as \emph{s-values}, the manifold predictor defines a set of $N$ isosurfaces that represent our static geometry:
\begin{equation}
    \mathcal{S}_i = \left\{ (x,y,z) \ | \ \cG(x,y,z) = s_i \right\}.
\end{equation}
In our appearance model, we first transform points on each manifold to UV-space coordinates via a fixed function $f : (x,y,z) \in \mathcal{S}_i \to (u,v) \in [-1,1] \times [-1,1] $. For each manifold $i \in \cI$ and each frame $j \in \cJ \triangleq \{1, 2, \dots, K\}$, a texture predictor $\cT_{ij}$ defines RGB and transparency fields:
\begin{equation}
    \cT_{ij} : (u, v) \in [-1,1]\times[-1,1] \to (r, g, b, \alpha) \in \R^4, \ \, i \in \cI \ \& \ j \in \cJ.
\end{equation}
Note that we do not estimate volume density as is the case for traditional volume rendering, but instead model 3D point transparency with an alpha value. This makes the radiance accumulation independent of the ray path, which is crucial for enabling the next step of exporting the learned manifolds as textured mesh layers. Our approach can be treated as a generalization of the multi-plane image representation~\cite{46965}, where we optimize arbitrary 2D surfaces instead of planes.

Once the training is completed, we collect samples across each manifold at a specific resolution and export these collections of 3D points, UV-coordinates, and RGBA values as a single set of topologized triangle meshes with UV-textures that can be efficiently rendered on legacy graphics renderers. We illustrate our training and inference pipelines in \autoref{fig:method}.

\subsection{Dataset}
We use the publicly available dataset Multiface~\cite{wuu2022multiface}, from which we gather multi-view videos of 3 subjects from V1 of the dataset (subject IDs \texttt{002914589}, \texttt{002643814}, \texttt{5372021}), and 2 subjects from V2 (subject IDs \texttt{002421669}, \texttt{002645310}). We pick $K = 60$ consecutive frames from each video sequence where subjects perform facial expressions freely. For all subjects, we scale and transform the scene parameters such that the subjects are centered at the origin and oriented along the positive $x$-axis with up-vector aligned with the positive $z$-axis, and that the camera centers are distributed roughly $1$ unit away from the subjects. We discard the cameras with elevation angles of more than $45^\circ\!$ and azimuth angles of more than $90^\circ\!$, which yields a set of cameras in the $x>0$ half-space. For each subject, we also hold out $2$ cameras to perform quantitative evaluations. This yields us $23$ training cameras for V1 subjects and $50$ training cameras for V2. Finally, we downsample all images to $768\times500$ resolution while adjusting the camera parameters accordingly. We do not perform any background masking as our method is able to separate the foreground significantly either by restricting the scene volume or by placing the background into the view-dependent component of radiance, which is discarded at inference time. 

\subsection{Model Architecture and Training}
Given $K$ frames from a multi-view video with corresponding camera parameters, we sample $M$ points uniformly along each camera ray and compute the intersections between these rays and the manifolds using the differentiable ray-manifold intersection algorithm~\cite{niemeyer2020differentiable} adopted in GRAM~\cite{deng2022gram}. In our method, we sample $M = 256$ points along each ray, set the number of manifolds to $N = 12$, and train our model using videos consisting of $K = 60$ frames for each subject.

Previous techniques that model dynamic scenes with radiance manifolds \cite{anifacegan,bakedavatar} have used explicit learned deformation of the manifolds to model scene animation. On the contrary, we model all frames of a dynamic sequence with a single set of static manifolds, which poses a non-trivial challenge. To achieve this, our technique uses a unique sequence of steps, where we 1) transform intersection points to UV-space, 2) separate RGB predictions into view-independent and view-dependent components, and 3) estimate the transparency of each intersection directly without computing volume densities. Given an intersection $\mathbf{p} = (x,y,z) \in \R^3$ and a fixed center $\mathbf{c} \in \R^3$ of a unit sphere, we first project the intersection to the surface of the sphere and obtain $\mathbf{p}' \triangleq (x', y', z') = (\mathbf{p} - \mathbf{c}) / \Vert \mathbf{p} - \mathbf{c} \Vert$. We then calculate the UV-space coordinates as $u = \frac{2}{\pi}\sin^{-1}(z')$ and $v = \frac{2}{\pi}\tan^{-1}(y' / x')$. The texture predictor receives UV-coordinates and the s-values of the intersections, as well as the frame index that is mapped to a learned latent code that conditions the predictions. The texture predictor is then branched into two layers that predict single-channel view-dependent compontent and three-channel view-independent component, former of which is conditioned on the view direction. The outputs of these branches are added together to produce the final RGB prediction. Such architecture allows us to discount view directional effects at inference time and achieve view-consistent rendering of exported meshes. Furthermore, it also helps us to separate most of the background from foreground by attributing the background to view-dependent component, particularly if the background is primarily grayscale, thus eliminating the need for explicit background removal. Finally, the alpha values are predicted as the raw output of our texture predictor, which can be directly used to alpha-composite our $N$-layered representation.

\subsubsection{Training details} To promote training stability, we adopt the manifold initialization technique~\cite{atzmon2020sal} used in  GRAM~\cite{deng2022gram} and begin training with sphere-like manifolds centered at $\mathbf{c}$. We optimize our model in an end-to-end fashion by adopting $\ell_1$ loss between the predicted and ground truth pixel values. To ensure that the appearance is mostly explained by the view-independent component, we penalize the output of the view-dependent branch with $\ell_2$ penalty. To promote more stability, we apply $\ell_2$ regularization to all manifold predictor layers except for the final one. The manifold and texture predictors are optimized jointly by minimizing the loss function
\begin{equation}
    \mathcal{L} = \mathcal{L}_{\mathrm{rec}} + \lambda_{\mathrm{vd}} \mathcal{L}_\mathrm{vd} + \lambda_{\mathrm{reg}} \mathcal{L}_\mathrm{reg}
\end{equation}
where $ \mathcal{L}_{\mathrm{rec}}$ is the $\ell_1$ reconstruction loss, $\mathcal{L}_\mathrm{vd}$ is the view-dependent branch penalty, and $\mathcal{L}_\mathrm{reg}$ is the manifold regularization with $\lambda_{\mathrm{vd}} = 1.0$ and $\lambda_{\mathrm{reg}} = 0.0001$. The manifold and texture predictors are jointly optimized using the Adam optimizer~\cite{adam} with initial learning rates of $0.0007$ and $0.0010$, and exponential decay rates of $0.05$ and $0.20$ per $200\,000$ iterations, respectively. Using a batch size of $32\,768$ rays sampled across all training frames and views, we perform training for $500\,000$ iterations for each subject.

\subsubsection{Architecture Details} The manifold predictor is implemented as an MLP with $3$ hidden layers of widths $128$ and a final layer, where we choose the set of fixed scalars $\{s_i\}_{i=1}^N$ so that the initial concentric and sphere-like surfaces roughly falls within $\pm 0.03$ units of the surface of the face. These scalars are tuned slightly for each subject according to the size of the faces inferred by the tracked meshes provided in the Multiface dataset~\cite{wuu2022multiface}. The texture predictor is implemented as an MLP with $8$ hidden layers of widths $256$ and $2$ final layers corresponding to view-independent and view-dependent branches. Both input points and view directions undergo positional encodings and the frame indices are mapped to $32$-dimensional learned latent codes through an embedding layer. Encoded input points and frame indices are fed into the MLP at its first layer whereas the encoded view directions are concatenated to the input to the view-dependent branch.

\begin{figure}[t]
  \centering
  \includegraphics[width=\linewidth]{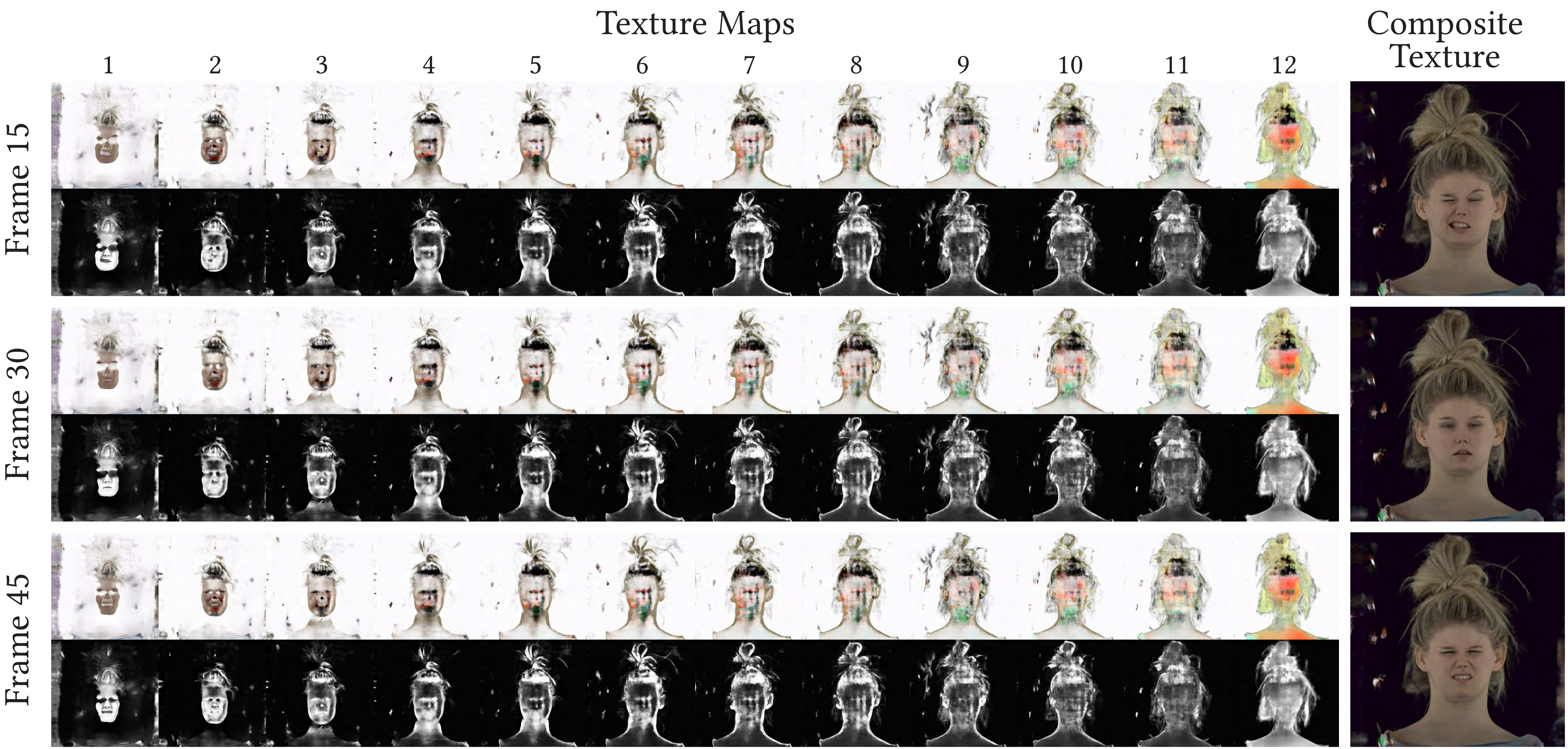}
  \caption{\textbf{Video texture visualization.} We illustrate $3$ frames ($15^\mathrm{th}$, $30^\mathrm{th}$, and $45^\mathrm{th}$ frames) from the learned texture video of subject \texttt{002914589}. For each frame, we show full RGBA and alpha-only UV-space texture maps in the top and bottom rows, respectively.}
  \label{fig:texture-viz}
\end{figure}

\subsection{Exporting Layered Meshes and Textures}
At test time, we gather points across the unit \emph{hemisphere} by collecting azimuth and elevation angles in $[-\pi /2 , \pi /2]$ uniformly at resolution $R\times R$ and shoot rays towards the sphere center $\mathbf{c}$. We set this center $0.25$ units away from the scene center in the direction of negative $x$-axis to ensure that the entire scene is encompassed by the hemisphere. This gives us $R\times R$ samples across each manifold with UV-coordinates that are distributed uniformly in $[-1,1]\times[-1,1]$. For each of the $K$ frames, these points are used to query the texture predictor to yield $N$ RGBA texture maps at resolution $R\times R$, where the view-dependent branch is bypassed to ensure the texture maps are fully diffuse. Our simple spherical projection yields reasonable texture mapping despite slight distortions near the edges of the maps~\cite{Du2018Montage4D}. We emphasize that we export the alpha channel as $8$-bit images and hence store them very efficiently without sacrificing the visual quality. Finally, while the texture maps vary according to their respective frame indices, the geometry is constant across all frames.

To export our manifold-based representation to explicit surfaces, we reconstruct meshes from each of the $N$ point clouds of size $R\times R$ via Poisson surface reconstruction~\cite{kazhdan2006poisson}, where the normals for each point are computed with respect to their neighboring points. We then simplify these meshes using a mesh decimation algorithm to reduce the number of vertices to a target mesh resolution $R^\mathrm{m}\times R^\mathrm{m}$. Finally, for each vertex in the simplified mesh, we determine the nearest point in the original point cloud and assign its corresponding UV-coordinate. The texture maps, on the other hand, can be downsampled to a specific target resolution $R^\mathrm{t}\times R^\mathrm{t}$ to meet the memory requirements of the renderer. To summarize, our final assets are: 1) a single set of $N$ triangle meshes, each with number of vertices less than the target resolution $R^\mathrm{m}\times R^\mathrm{m}$ and 2) $K$ sets of $N$ RGBA texture maps at resolution $R^\mathrm{t}\times R^\mathrm{t}$ that form a UV texture video. We illustrate $3$ frames from an example texture video along with composited texture maps in \autoref{fig:texture-viz}.

\subsection{Rendering on Game Engines}
\begin{figure}[t]
  \centering
  \includegraphics[width=\linewidth]{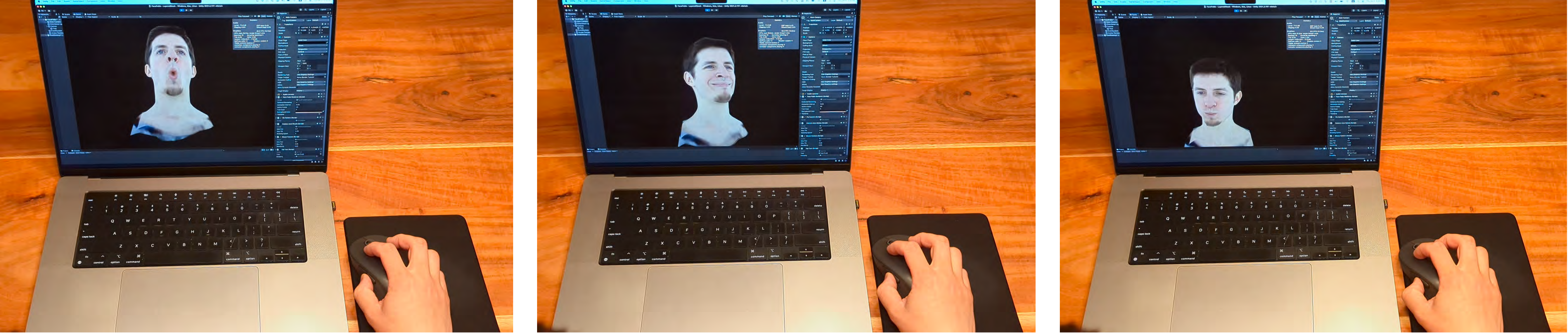}
  \caption{\textbf{Free-viewpoint rendering on Unity.} Our representation allows for free-viewpoint rendering of dynamic 3D volumes on consumer hardware. Please refer to the supplementary material for the videos.}
  \label{fig:unity}
\end{figure}

Our rendering pipeline in Unity using the exported layered mesh and texture sequences runs in real-time. We leverage two-pass deferred shading~\cite{Deering1988Triangle} on the GPU. When given a camera pose at runtime, we generate $N$ G-buffers by shading each mesh layer and its opacity in a single render pass. Modern game engines use multiple render targets (MRT) for this purpose and we used culling masks to achieve this in Unity.

For a small number of layers (\textit{e.g.}, $N\!<\!16$, which is the maximum texture sampler count supported in Unity), we composite G-buffers on the GPU by tracing a ray through all layers in one pass, similar to accumulating luminance in the traditional volume rendering pipeline. For more than 16 layers, we suggest using a prefix sum algorithm~\cite{blelloch1990prefix} on the GPU for efficient layer compositing.

In our experiments with $N = 12$, we achieved real-time performance on a 2019 Macbook Pro with an M1 Max chip and Unity 2021.3. This was consistent across five datasets and over 1,000 frames. The average rendering time per frame was under 17 ms (above 60 FPS) at a rendering resolution of $2560\times1440$, even for our largest reconstructed mesh of 6.3M triangles. 

%% file: sections/4_results.tex
\section{Experiments and Results}
We evaluate the performance of our method on several subjects from the Multiface dataset~\cite{wuu2022multiface}, where we provide qualitative and quantitative comparisons against state-of-the-art neural rendering methods. We then perform more analysis of the configurability of our representation by assessing its performance with respect to varying number of manifolds, mesh resolution, and texture resolution. 

\subsection{Qualitative Results}
\begin{figure}[t]
  \centering
  \includegraphics[width=\linewidth]{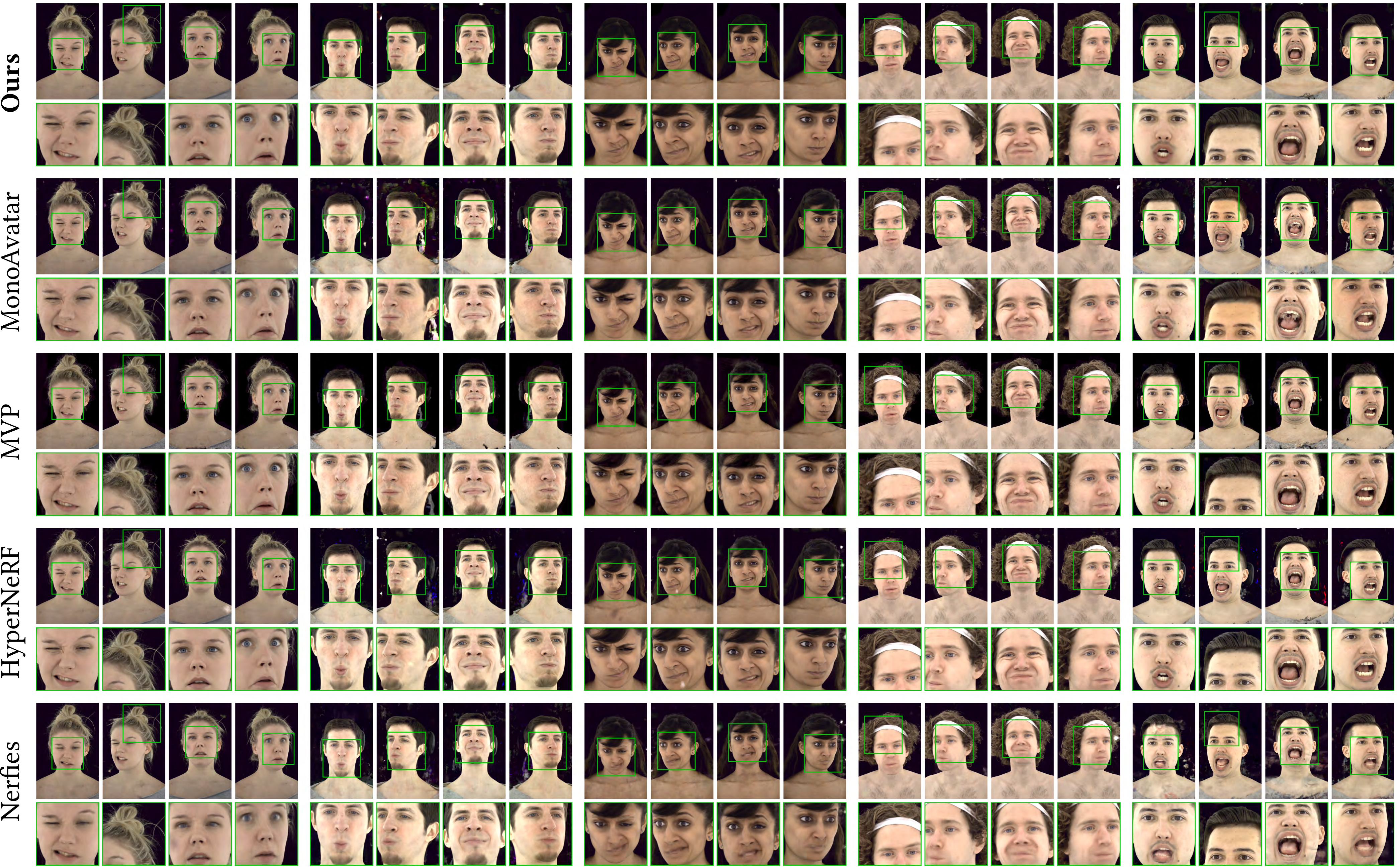}
  \caption{\textbf{Qualitative comparisons.} Our method achieves comparable visual quality to state-of-the-art neural rendering techniques while  facilitating very efficient rendering of dynamic sequences on legacy graphics software without any custom integration of ML pipelines.}
  \label{fig:comparisons}
\end{figure}
We train our pipeline individually on $5$ multi-view video sequences from~\cite{wuu2022multiface}, and illustrate our novel view synthesis results in~\autoref{fig:comparisons}. Here, we provide comparisons against $4$ state-of-the-art neural rendering methods---MonoAvatar~\cite{monoAvatar}, MVP~\cite{mvp}, HyperNeRF~\cite{park2021hypernerf}, and Nerfies~\cite{park2021nerfies}. Despite discretizing the 3D volume into only $N = 12$ manifolds and hence sampling much fewer points across the scene during both training and evaluation, our approach manifests a comparable performance against other techniques. Furthermore, our technique does not require any MLP query or a custom pipeline during rendering and thus can be exported into a game engine, where we can perform free-viewpoint rendering of a dynamic 3D scene. We import our layered meshes and UV-textures into Unity and achieve the results demonstrated in~\autoref{fig:unity}. We encourage the reader to refer to the supplementary material for video visualizations. 

\begin{figure}[t]
  \centering
  \includegraphics[width=\linewidth]{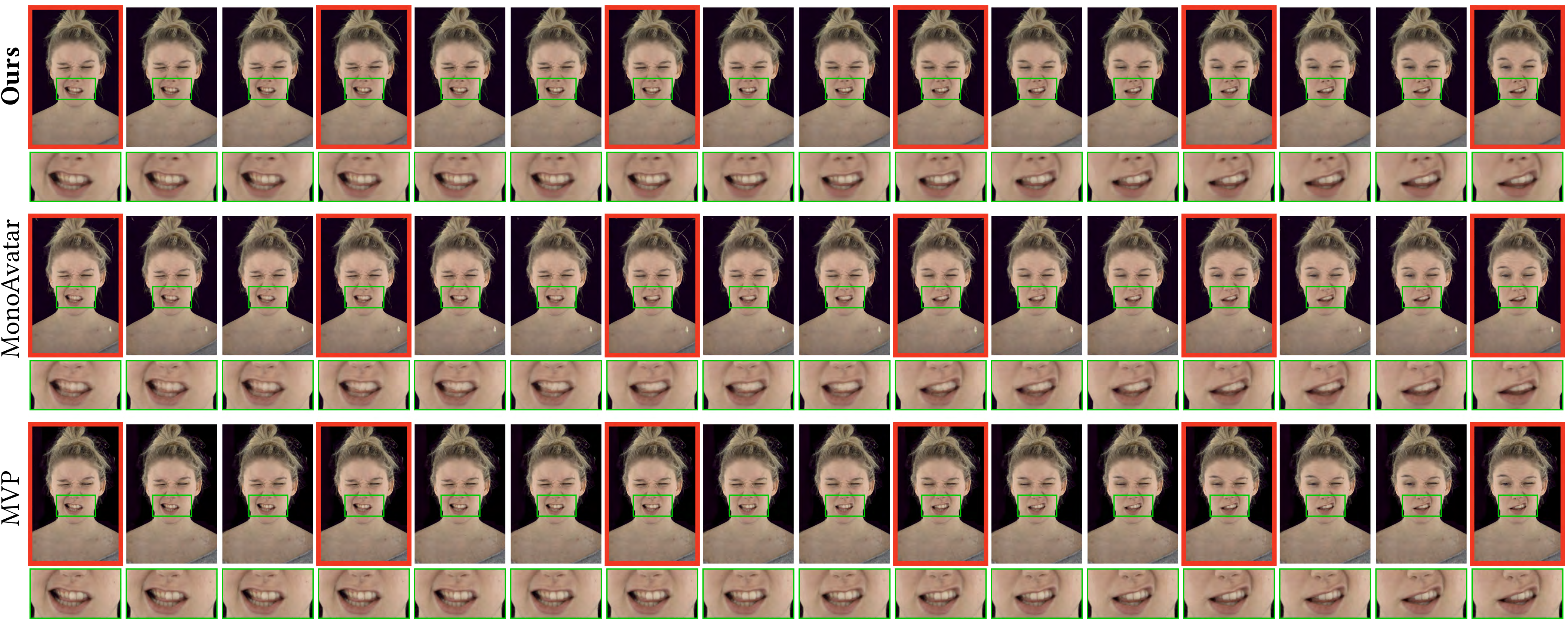}
  \caption{\textbf{Frame interpolation results.} Interpolating between the learned latent codes of frame indices allows us to achieve high-quality temporal interpolation between training frames with comparable performance to other approaches. Original frames are highligted in red.}
  \label{fig:interp}
\end{figure}

By interpolating between the learned latent codes of different frames at inference time, we can render our representation at higher frame rates to enhance the overall visual quality. We depict our frame interpolation results and provide comparisons in~\autoref{fig:interp}, where we observe comparable performance against other methods. Please refer to the supplementary material for the videos.

\subsection{Quantitative Results}
For each subject, we perform quantitative evaluations on $2$ held-out cameras across all $K = 60$ frames, totaling $120$ images. In~\autoref{table:comparisons}, we report average image quality metrics for our method and other methods in PSNR, SSIM~\cite{ssim}, and LPIPS~\cite{lpips}, where we consistently observe comparable performance across all methods. We also report VRAM usage, required disk storage, and frame rates for each of the methods, where we compress individual texture maps into a video and apply mesh compression to individual meshes using Draco\footnote{\url{https://google.github.io/draco/}} with no quantization of vertex positions and texture coordinates, and using the lowest compression amount. From our results, we observe that our method is able to achieve higher frame rates despite utilizing a comparable amount of storage against other methods. Note here that other methods can be run with much lower VRAM usage by reducing the batch size down to single ray per batch. 

All ML training, including ours and the state-of-the-art methods, was done on a workstation with NVIDIA V100 GPU. Since the state-of-the-art methods require a Linux workstation with NVIDIA GPU also for inference, they were evaluated and profiled on this same workstation. Our method does not require such special ML integration, and we perform our evaluation on the Unity game engine on a 2019 Macbook Pro laptop.

\begin{table}
\small
  \caption{\textbf{Quantitative comparisons.} Our method attains comparable visual quality across various metrics while utilizing significantly less VRAM and enabling much higher frame rates. The image quality metrics are averaged over a total of $600$ test images of all $5$ subjects.}
  \label{table:comparisons}
  \begin{tabular}{ccccccc}
    \toprule
    Method & PSNR\,$\uparrow$ & SSIM\,$\uparrow$ & LPIPS\,$\downarrow$ & VRAM$\,\downarrow$ & Disk\,$\downarrow$ & FPS\,$\uparrow$ \\
    \midrule
    Ours & $25.49 \pm 3.16$ &  $0.788 \pm 0.069$ & $0.356 \pm 0.038$ & $602\,\mathrm{MiB}$ & $118\,\mathrm{MiB}$ & $>\!60$\\
    MonoAvatar~\cite{monoAvatar} & $24.59  \pm 1.71$ & $0.786 \pm 0.042$ & $0.341 \pm 0.018$ & $2746\,\mathrm{MiB}$ & $296\,\mathrm{MiB}$ & $0.33$\\
    MVP~\cite{mvp}  & $26.20 \pm 2.34 $ & $0.738 \pm 0.54$ & $0.313\pm 0.025$ & $1412\,\mathrm{MiB}$ & $356\,\mathrm{MiB}$ & $12.7$\\
    HyperNeRF~\cite{park2021hypernerf}  & $26.91 \pm 3.14$ & $0.845 \pm 0.0394$ & $0.305 \pm 0.041$ & $2861\,\mathrm{MiB}$ & $15\,\mathrm{MiB}$ & $0.02$\\
    Nerfies~\cite{park2021nerfies} & $26.11 \pm 3.15$ & $0.815 \pm 0.042$ & $0.332 \pm 0.044$ & $4205\,\mathrm{MiB}$ & $15\,\mathrm{MiB}$ & $0.03$ \\
  \bottomrule
\end{tabular}
\end{table}

\subsection{Ablation Studies}
\begin{figure}[t]
  \centering
  \includegraphics[width=\linewidth]{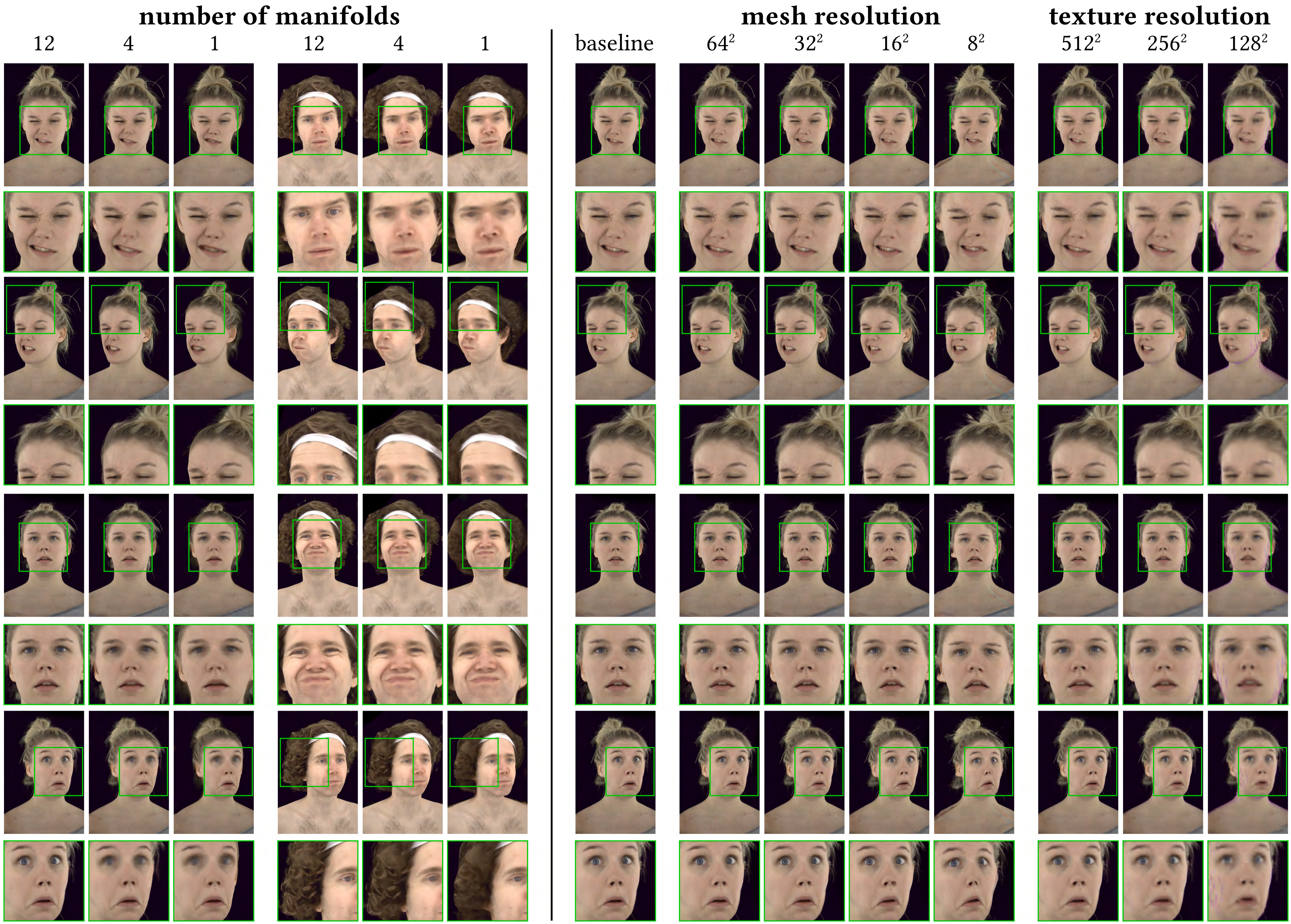}
  \caption{\textbf{Ablation results.} \textit{Number of manifolds.} Using a sufficient number of manifolds is essential to attain photorealism and volumetric effects. \textit{Mesh resolution.} We can decimate the exported meshes to much lower resolutions without sacrificing significant visual quality. \textit{Texture resolution.} We can modify texture resolution arbitrarily at inference time to trade off image quality with rendering efficiency as desired.}
  \Description{Ablation results.}
  \label{fig:ablation}
\end{figure}

Since radiance manifolds~\cite{deng2022gram} constrain 3D volumes to a number of implicit surfaces, the rendering quality is strongly influenced by the number of manifolds chosen before training. We provide qualitative and quantitative comparisons across different numbers of manifolds in~\autoref{fig:ablation} and~\autoref{tab:num-manifolds}. Here, we observe that not only the rendering quality suffers with decreasing number of manifolds, capturing volumetric effects also requires a sufficient number of samples.

\begin{table}
\small
  \caption{\textbf{Ablation on number of manifolds.} While significant gains in memory efficiency can be achieved by reducing the number of manifolds, it has a noteable effect on the visual quality. Numbers are averaged over a total of $240$ test images of two subjects with IDs \texttt{002914589} and \texttt{002421669}. We report total disk storage required by the meshes and the video texture as well as the total number of triangles in all meshes.}
  \label{tab:num-manifolds}
  \begin{tabular}{cccccc}
    \toprule
    Num.\ Manifolds & PSNR\,$\uparrow$ & SSIM\,$\uparrow$ & LPIPS\,$\downarrow$ & Num.\! Tri.\,$\downarrow$ & Disk\,$\downarrow$\\
    \midrule
    $12$ & $25.54$ & $0.760$ & $0.341$  & $6264412$ & $125.1\,\mathrm{MB}$ \\
    $4$ & $24.47$ & $0.728$  & $0.372$ & $2088497$  & $42.9\,\mathrm{MB} $ \\
    $1$ & $22.51$ & $0.704$ & $0.395$ & $522242$  & $11.0\,\mathrm{MB} $\\
  \bottomrule
\end{tabular}
\end{table}

Our exported representation allows for trading off image quality with memory efficiency by performing standard operations such as mesh simplification and texture downsampling. After reconstructing a single set of meshes via Poisson surface reconstruction~\cite{kazhdan2006poisson}, we decimate each of these meshes to meet a target number of vertices.  We illustrate qualitative and quantitative evaluations for varying mesh resolutions in~\autoref{fig:ablation} and~\autoref{tab:mesh-res}. We observe that the image quality does not undergo a significant drop until $16\times16$ resolution per surface. This is because our layered mesh representation does not manifest high-frequency changes while still allowing for state-of-the-art rendering quality via learned alpha-manifolds. This provides us with an extremely lightweight geometry representation without sacrificing any visual quality or volumetric effects.

The texture resolution, on the other hand, naturally plays a vital role in rendering quality. To compare, we individually subsample each of the texture maps across all manifolds and frames bilinearly, and render each frame at original training resolution $768\times500$. We illustrate our results in~\autoref{fig:ablation} and~\autoref{tab:texture-res} where we observe a notable reduction in quality at $128\times128$ resolution.

\begin{table}
\small
  \caption{\textbf{Ablation on mesh resolution.} Despite reducing the memory footprint of the geometry, visual quality is maintained for resolutions as low as $32\times32$. For all mesh resolutions, the texture resolution is set to $1024\times1024$ and requires $5.9$ MB storage after video compression. Numbers are averaged over $120$ test images of a single subject with ID \texttt{002914589}.}
  \label{tab:mesh-res}
  \begin{tabular}{cccccc}
    \toprule
    Mesh resolution & PSNR\,$\uparrow$ & SSIM\,$\uparrow$ & LPIPS\,$\downarrow $& Num.\! Tri.\,$\downarrow$ & Disk (mesh)\,$\downarrow$\\
    \midrule
    $512\times 512$ & $30.10$ & $0.858$ & $0.284$  & $6261920$  & $118.1\,\mathrm{MB}$ \\
    $256\times 256$ & $29.58$ & $0.838$ & $0.293$  & $1377119$  & $22.2\,\mathrm{MB}$\\
    $128\times 128$ & $29.57$ & $0.839$ & $0.290$  & $250740$  & $4.1\,\mathrm{MB}$\\
    $64\times 64$ & $29.49$ & $0.837$ & $0.289$  & $43224$ & $721\,\mathrm{KB}$\\
    $32\times 32$ & $29.12$ & $0.831$ & $0.291$  & $9962$ & $170\,\mathrm{KB}$\\
    $16\times 16$ & $27.26$ & $0.795$ & $0.306$  & $2574$ & $38\,\mathrm{KB}$\\
    $8\times 8$ & $23.81$ & $0.705$ & $0.364$  & $728$ & $11\,\mathrm{KB}$\\
  \bottomrule
\end{tabular}
\end{table}

\begin{table}
\small
  \caption{\textbf{Ablation on texture resolution.} We can reduce the memory footprint of the video texture by simply subsampling each frame at inference time. For renders of resolution $768\times500$, a noteable drop in quality occurs at $256\times256$ texture resolution. For all texture resolutions, the mesh resolution is set to $512\times512$, hence the number of triangles and disk storage for meshes are constant and are $6261920$ and $118.1$ MB, respectively. Numbers are averaged over $120$ test images of a single subject with ID \texttt{002914589}. Note that the texture video size will increase with the number of frames in the input video.}
  \label{tab:texture-res}
  \begin{tabular}{ccccc}
    \toprule
    Texture resolution & PSNR\,$\uparrow$ & SSIM\,$\uparrow$ & LPIPS\,$\downarrow$ & Disk (texture)\,$\downarrow$\\
    \midrule
    $1024\times 1024$ & $30.10$ & $0.858$ & $0.284$  &  $5.9\,\mathrm{MB}$ \\
    $512\times 512$ & $30.29$ & $0.859$  & $0.303$  & $2.1\,\mathrm{MB}$ \\
    $256\times 256$ & $29.51$ & $0.836$ & $0.341$  & $667\,\mathrm{KB}$\\
    $128\times 128$ & $27.32$ & $0.791$ & $0.404$   & $183\,\mathrm{KB}$\\
  \bottomrule
\end{tabular}
\end{table}

%% file: sections/5_future_work.tex
\section{Limitations and Future Work}

\begin{figure}[t]
  \centering
  \includegraphics[width=\linewidth]{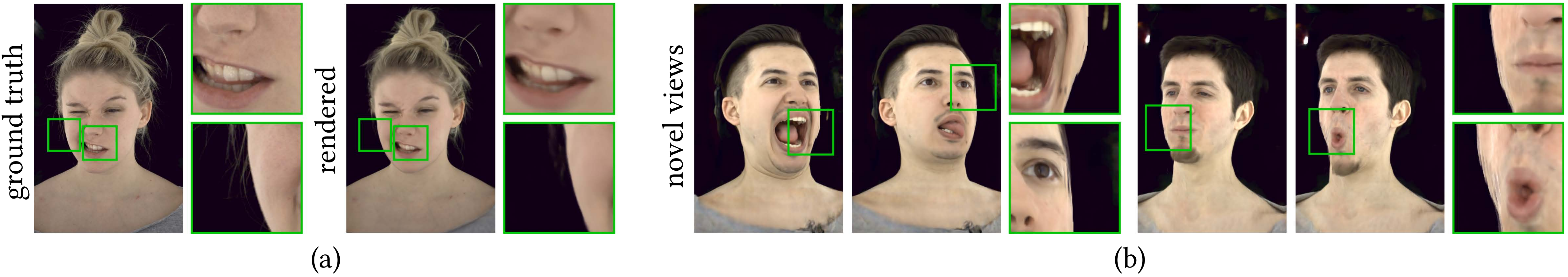}
  \caption{\textbf{Limitations.} While our method achieves good image quality in general, it suffers from some drawbacks. (a) Since we export only view-independent radiance to the texture, we cannot render specularities such as the ones on the nose, teeth and cheeks. (b) At extreme out-of-training-distribution viewpoints, our method sometimes exhibits shell artifacts due to transparency of the layered mesh from grazing views. Please refer to the supplementary material for video visualizations.}
  \label{fig:limitations}
\end{figure}

Since we export view-independent texture maps, our results do not exhibit view directional effects, as shown in~\autoref{fig:limitations}(a). Our pipeline could be extended to estimate specular or roughness maps from the view-dependent component to enable plausible specular relighting in graphics pipelines.
While we demonstrate that the radiance manifolds can be leveraged to export lightweight geometry and appearance models, joint learning of these models poses challenges in training stability and requires careful tuning of relative learning rates of the two models. Besides these challenges, we observe that sampling across discrete manifolds instead of the entire 3D volume causes shell artifacts in extreme poses, as illustrated in~\autoref{fig:limitations}(b). Our experiments suggest that these artifacts can be mitigated by initializing the manifolds according to the size of the scene and keeping the distance between the consecutive manifolds sufficiently small, while also ensuring that these distances are large enough to allow for volumetric effects. In addition to these heuristics and the regularization of the manifold predictor weights, more sophisticated regularization techniques~\cite{gropp2020implicit} can be utilized to promote training stability and improve geometry predictions.

We should note that our method involves sampling across the 3D volume by casting a single ray per pixel and query MLPs for each point across these rays, which is prone to producing aliasing artifacts~\cite{barron2022mipnerf360} and lead to slow training~\cite{mueller2022instant}. Using more recent neural rendering frameworks that, for example, combine anti-aliasing techniques and fast grid-based representations~\cite{barron2023zipnerf} would be a possible next step towards improving the overall performance and visual quality of our method, although incorporating radiance manifolds into such pipelines can present nontrivial challenges.

While our spherical mapping to UV-space coordinates works well since our learned radiance manifolds are smooth and mostly convex, such mapping could be problematic for non-convex regions such as the nose. Also, our spherical sampling technique at inference time could be modified to have denser number of samples around more detailed regions such as the eyes and the hair. A non-linear UV mapping technique for the face in the context of radiance manifolds would be an exciting future research exploration.

%% file: sections/6_conclusion.tex
\section{Conclusion}
In this work, we introduce FaceFolds, a novel representation for high-quality and memory-efficient volumetric rendering of dynamic facial performances in legacy renderers. We achieve this by leveraging radiance manifolds to model the animated performance. Our novel contribution includes a unique sequence of operations and design choices required to make the radiance manifold framework view-\textit{independent} to enable exporting of the layered mesh and video textures. Once these assets are obtained, our representation does not require any ML-based operations or complex computations, and hence can be easily rendered in standard graphics software on consumer hardware at high frame rates. Our results demonstrate that we still achieve state-of-the-art rendering quality despite securing notable gains in memory and compute footprint.